# MONO-HYDRA: REAL-TIME 3D SCENE GRAPH CONSTRUCTION FROM MONOCULAR CAMERA INPUT WITH IMU

U.V.B.L. Udugama [1*], G. Vosselman [1], F. Nex [1]

[1] Faculty of Geo-Information Science and Earth Observation (ITC), University of Twente, Enschede, The Netherlands
(b.udugama, george.vosselman, f.nex)@utwente.nl

**ABSTRACT:**

The ability of robots to autonomously navigate through 3D environments depends on their comprehension of spatial concepts, ranging from low-level geometry to high-level semantics, such as objects, places, and buildings. To enable such comprehension, 3D scene graphs have emerged as a robust tool for representing the environment as a layered graph of concepts and their relationships. However, building these representations using monocular vision systems in real-time remains a difficult task that has not been explored in depth.

This paper puts forth a real-time spatial perception system Mono-Hydra, combining a monocular camera and an IMU sensor setup, focusing on indoor scenarios. However, the proposed approach is adaptable to outdoor applications, offering flexibility in its potential uses. The system employs a suite of deep learning algorithms to derive depth and semantics. It uses a robocentric visual-inertial odometry (VIO) algorithm based on square-root information, thereby ensuring consistent visual odometry with an IMU and a monocular camera. This system achieves sub-20 cm error in real-time processing at 15 fps, enabling real-time 3D scene graph construction using a laptop GPU (NVIDIA 3080). This enhances decision-making efficiency and effectiveness in simple camera setups, augmenting robotic system agility. We make Mono-Hydra publicly available at: https://github.com/UAV-Centre-ITC/Mono_Hydra.

## 1. INTRODUCTION

Real-time high-level representations of environments are crucial for robots and autonomous systems, enabling efficient comprehension and execution of human instructions, fast planning, and comprehensive situational understanding. These representations can revolutionise various applications, including search and rescue, warehouse monitoring, maintenance, and surveillance. By replicating human-level understanding, robots equipped with these representations can excel in tasks such as identifying survivors and hazards, optimising inventory management, conducting efficient maintenance operations, and enhancing surveillance capabilities, all of which depend on their comprehensive understanding of the situation. In recent years, 3D scene graphs have emerged as powerful high-level representations of environments (Hughes et al., 2022), yet real-time construction remains a significant challenge. While some works allow real-time operation, they are restricted to RGB-D or Lidar-like sensor systems where depth perception is readily available (Bavle et al., 2022).

The exploration of monocular camera setups for achieving real-time 3D scene graph generation remains an understudied area in contrast to the existing techniques utilising RGB-D setups or 3D LIDARs. This research gap is significant considering that monocular cameras offer advantages in compactness and agility compared to the aforementioned sensor setups. The utilisation of monocular cameras shows immense promise, especially in applications like UAV operations where minimising payload is crucial for prolonged missions. However, to the best of our knowledge, there is limited work on real-time 3D scene graph generation from monocular camera input with an IMU.

This paper presents an approach that leverages a set of deep neural networks combined with an IMU to enhance visual odometry and enable the real-time construction of a 3D scene graph, as in Figure 1. To accomplish this, we adopt the Hydra framework (Hughes et al., 2022), specifically designed to integrate semantic mesh and odometry data towards building scene graphs. Our proposed suite of deep learning algorithms is designed to generate a real-time 3D semantic mesh, providing the essential inputs for constructing the scene graph.

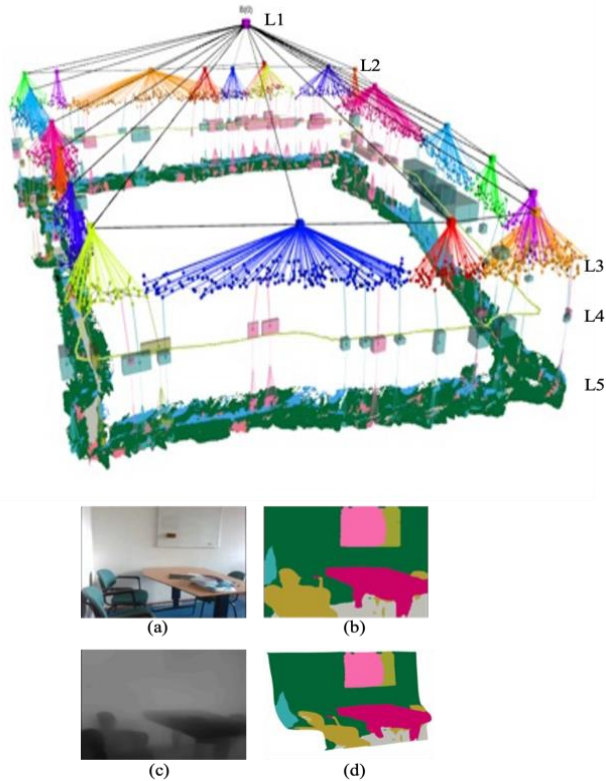

**Figure 1.** Real-time 3d scene graph generated using the Mono-Hydra framework on the 2nd floor of the ITC building. (a) RGB image (b) Predicted Semantics (c) Predicted depth (d) 3D semantic mesh. Levels in the graph: L1- buildings, L2- Rooms, L3-Places, L4-Objects, L5-Metric-semantic mesh.

---
* Corresponding author

## 2. RELATED WORKS

### 2.1 Spatial Perception Systems and 3D Scene Graphs

In recent years, 3D scene understanding has become a crucial problem in robotics, computer vision, and augmented reality. With the advent of deep learning, there have been significant advancements in 3D reconstruction and recognition, enabling the development of real-time 3D scene graphs. 3D scene graphs are a recent development and serve as a powerful way of representing 3D environments (Armeni et al., 2019). They are created as a layered graph where each node represents a spatial concept at different levels of abstraction, ranging from low-level geometry to high-level semantics such as objects, places, rooms, and buildings. The connections between these nodes are represented as edges, which convey the relations between different concepts. These 3D scene graphs can act as an advanced "mental model" for robots, allowing them to understand better and navigate complex environments (Ravichandran et al., 2022).

Spatial perception systems use modern machine learning techniques to extract semantics from visual feeds and hierarchical semantic representations like 3D scene graphs to build an environmental model. Kimera (Rosinol et al., 2021) and Hydra (Hughes et al., 2022) are relevant approaches to studying real-time performance and adapting hierarchical representations for robot spatial perception systems. Kimera provides a novel metric-semantic hierarchical representation of the environment and practical algorithms to infer it from data using a 3D Dynamic Scene Graph (DSG) (Armeni et al., 2019) representation for actionable spatial perception. It is worth noting that the generation of the DSG is not real-time in Kimera; although the creation of the metric-semantic reconstruction happens in real-time, the rest of the scene graph is built at the end of the run and requires a few minutes to parse the entire scene (Rosinol et al., 2021). The study demonstrates the potential queries that can be implemented on a DSG and shows how a robot can use a DSG to understand and execute high-level instructions. Hydra is another relevant approach designed to overcome the main drawback of the Kimera framework, which is the real-time performance, and it has shown promising results in building and maintaining a semantic map of the environment that can be used for localisation and planning tasks.

Another recent approach, Situational Graphs called S-Graphs+ (Bavle et al., 2022), combines a SLAM graph with a 3D scene graph to model the environment. S-Graphs+ is a four-layered factor graph optimised in real-time to estimate the robot's pose and map while leveraging high-level information about the environment. The paper introduces new room and floor segmentation algorithms utilising mapped wall planes and free-space clusters. Still, it depends on the 3d lidar input compared to the previously mentioned Hydra and Kimera frameworks, which use RGB-D camera inputs.

### 2.2 Monocular Depth Prediction

Single image depth estimation is challenging due to its ill-posed nature, but deep learning has successfully addressed this problem (Wu et al., 2021). Two main categories of methods include supervised depth estimation, which uses ground-truth depth maps for training, and self-supervised depth estimation, which does not require annotated data. Self-supervised methods include novel view synthesis and monocular video frames, and some approaches introduce additional constraints or multi-task learning to improve accuracy. Recent work has shown that enhancing loss functions can also lead to competitive results in monocular depth estimation, as demonstrated in the Monodepth2 (Godard et al., 2018) method.

Recently, there has been a lot of research interest in self-supervised depth estimation, mainly using left-right consistency. However, well-known works such as MonoDepth, MonoDepth2 (Godard et al., 2018), DepthHints (Godard et al., 2016), and LiteMono (Zhang et al., 2022) mostly concentrate on driving scenes and are trained on extensive driving datasets like KITTI and Cityscapes. It is not clear how these methods can be utilised in indoor environments. Learning depth in indoor environments using self-supervision is more challenging due to several factors. Firstly, indoor scenes have weaker structure priors than driving scenes, as objects can be cluttered and arranged arbitrarily. Secondly, indoor depth distribution can be concentrated in either near or far ranges, making it challenging to predict accurate metric depth. Thirdly, depth-sensing devices can move in 6DoF for indoor captures, making it necessary for networks to be more robust to arbitrary camera poses and complex scene structures. Lastly, large untextured regions, such as walls, make the commonly used photometric loss ambiguous. While DistDepth (Wu et al., 2021) and ZoeDepth architectures (Farooq et al., 2023) excel in predicting metric-accurate depth for unseen indoor scenes, their reliance on high-quality training data and limitations in handling challenging conditions such as low-light environments and occlusion may impact their performance and generalisability.

### 2.3 Monocular VIO

Visual Inertial Odometry (VIO) aims to estimate the poses of a sensing platform in unknown environments. Most approaches solve this problem from a global perspective by choosing a fixed global reference frame aligned with gravity. This approach is known as world-centric VIO but suffers from the observability mismatch issue between original and linearised systems (Rosinol et al., 2021; Z. Wang et al., 2022). Various remedies have been proposed, but they often trade off accuracy or efficiency. In contrast, a robocentric RVIO2 framework (Huai & Huang, 2022) reformulates the problem locally, where the body frame of the robot can be used as the instantaneous navigation frame of reference. The relative pose between every two locations of the robot is estimated, and the current pose with respect to the start (body) frame can always be recovered by incrementally merging new relative pose estimates. The observability mismatch issue, where insufficient sensor measurements to accurately estimate the system's state, does not exist for this approach, fundamentally improving the VIO estimator's consistency.

## 3. METHODOLOGY

This section describes our process for achieving real-time spatial perception with a sensor system occupied by a monocular camera and an IMU. The process starts by collecting real-time data from the monocular camera and the IMU, followed by pre-processing the data. Then, deep learning networks are utilised to estimate depth, perform semantic segmentation, and calculate visual-inertial odometry, as illustrated in Figure 2. Predicted depth information and semantic segmentation from the monocular camera are used to create a semantic mesh in real-time, which, along with the visual-inertial odometry data, are fed into Hydra (Hughes et al., 2022) to generate a 3D scene graph. In order to achieve real-time spatial perception and meet efficiency requirements, resources are divided and optimised for processing. The CPU (8 threads) is used for Hydra, while the GPU is used for deep learning networks. An Alienware m17 laptop with an Nvidia RTX 3080 GPU is utilised to run mono-hydra in a real-world setting using monocular RGB data

(848x480 – 30 fps) and IMU data (200 Hz) streams from a Realsense D435i sensor. This approach effectively achieves real-time spatial perception by combining the strengths of deep learning networks and visual-inertial odometry.

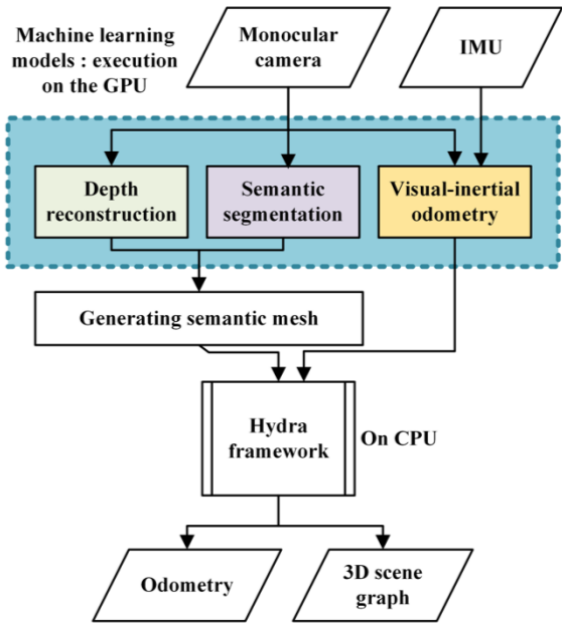

**Figure 2**. Processes in the Mono-Hydra framework.

### 3.1 Generating Semantic Mesh

The methodology involves the robotic operating system-based (ROS) package called depth_image_proc, which generates a 3D mesh from predicted depth and segmentation data. This process involves using a nodelet, which combines a registered depth image and an RGB image into an XYZRGB point cloud. In this use case, the semantic prediction will be used instead of the RGB image to colour the 3D mesh with semantic predictions.

**3.1.1 Monocular depth estimation:** Two state-of-the-art methods for monocular depth prediction were employed to evaluate the effectiveness of specialised architectures. The evaluation focused on predicting indoor metric-accurate depth and an outdoor relative depth predictor where scale had to be calculated by finding the average scale shift comparison between ground truth and predicted relative depth data (Steenbeek et al., 2022). These self-supervised networks were fine-tuned on the NYUv2 indoor dataset.

**DistDepth** (Wu et al., 2021) is a novel self-supervised learning approach that improves depth accuracy using an off-the-shelf relative depth estimator called DPT (Ranftl et al., 2021). DPT produces structured but only relative depth values that reflect depth-ordering relations but are metric-agnostic. To blend depth-ordering relations into metric depth estimation, DistDepth uses a structure distillation strategy that statistically and spatially promotes depth structural similarity. DistDepth only requires stereo image inputs without depth annotations and can predict structured and metric-accurate depth for unseen indoor scenes. Additionally, distillation helps to reduce DPT's large vision transformer to a smaller architecture, allowing for real-time inference on portable devices.

**LiteMono** (Zhang et al., 2022) is a lightweight architecture designed to extract effective features from input images using a lightweight encoder. The proposed architecture includes an encoder-decoder DepthNet (Anunay et al., 2021) and a PoseNet (Kendall et al., 2015) that estimate multi-scale inverse depth maps and camera motion between adjacent frames, respectively. The architecture also uses consecutive dilated convolutions (CDC) to enhance local features and a Local-Global Features Interaction (LGFI) module to model long-range information efficiently. By adopting cross-covariance attention instead of self-attention, the LGFI module reduces both memory and time complexity, making it suitable for lightweight models.

**3.1.2 Semantic segmentation:** The pre-trained model from the MIT Scene Parsing challenge (B. Zhou, 2016) is used for 2D semantic segmentation with HRNet (J. Wang et al., 2021). Few networks had pre-trained semantic segmentation models for ADE20k (Zhou et al., 2017a) data set with 20210 training image sets and 2000 validation sets, where 150 classes were annotated. It is compatible with the inference toolchain that existed in Hydra for semantic class mapping, despite the availability of newer and more performant options. Hence HRNetV2 is selected despite having low FPS performance but higher pixel percentage accuracy for generating the most accurate 3d semantic meshes compared to newer models, as demonstrated in Table 1.

| Architecture | MIoU | Pixel Accuracy (%) | Inference Speed (fps) |
|---|---|---|---|
| MobileNetV2dilated+ C1_deepsup | 34.84 | 75.75 | 17.2 |
| ResNet18dilated+ C1_deepsup | 33.82 | 76.05 | 13.9 |
| UperNet50 | 40.44 | 79.80 | 8.4 |
| **HRNetV2** | **42.03** | **80.77** | **5.8** |

**Table 1.** Pretrained network architectures are available for the ADE20k data set under CSAILVison public repository without Multiscale testing (Zhou et al., 2020). Architectures like MobileNetV2dilated + C1_deepsup represent the encoder and decoder of the architecture, respectively. Inference speed is benchmarked on a single NVIDIA Pascal Titan-Xp GPU.

Furthermore, the output of the HRNet configured for the ADE20K dataset has 150 segmentation classes which need to be configured to be 20 classes as per the initial Hydra framework configuration, which we kept as it is. So, the output classes were truncated to 20 using a simple mapping. This is demonstrated in the example configuration below, which shows the mapping of the new truncated class number 12, 'chair', from the original 150 class set (with the corresponding names mentioned in the brackets)

*class_info/12/labels:*

- 19 (chair)          - 75 (swivel, chair)
- 30 (armchair)       - 110 (stool)
- 31 (seat)

### 3.2 Monocular VIO

Kimera VIO (Rosinol, Abate, et al., 2020) has been proposed along with the Hydra framework for performing experiments with stereo sequences. Even though Kimera VIO supports monocular image sequences, it is not robust and often gets lost with fast rotations. Hence monocular image sequences and IMU-based RVIO2 (Huai & Huang, 2022) framework are modified to interface with the Hydra framework. This paper uses a novel information-based estimator called R-VIO2 to improve efficiency and robustness for resource-constrained applications. The estimator's numerical stability and computational efficiency are significantly enhanced by utilising a square-root expression and an incremental QR-based update combined with back substitution. Additionally, joint online calibration of spatial transformation and time offset between visual and inertial sensors is employed to increase the estimator's robustness in the presence of unknown parameter errors.

### 3.3 Generating 3D Scene Graph

Hydra framework is developed in a way that provides 3D semantic mesh and odometry data to build and optimise 3D scene graphs. Hydra can be deployed in CPU (8 threads), so deep learning networks can be easily deployed into GPU. RVIO2 can use the rest of the CPU for its optimisations, as illustrated in Figure 3.

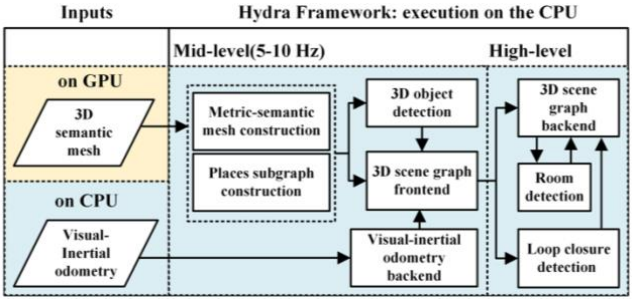

**Figure 3.** To link Hydra with 3D semantic mesh and VIO, HRNet and DistDepth/Lite-Mono networks were implemented on the GPU using CUDA, while VIO and Hydra (Hughes et al., 2022) were implemented on the CPU.

## 4. EXPERIMENTS

This section demonstrates that Mono-Hydra can construct 3D scene graphs in real-time with a level of precision comparable to other methods. Additionally, a comparison is made between the metric accuracy of the resulting semantic mesh and that of different approaches. Fine-tuning of the depth and semantic networks performed on Nvidia TITAN Xp GPU (12GB) and Mono-Hydra was tested on an Alienware m17 laptop with Nvidia RTX 3080 GPU (8GB).

### 4.1 Experimental Setup

**4.1.1 Datasets:** To test Mono-Hydra, we conducted experiments using a synthetic dataset called uHumans2 (uH2) (Rosinol et al., 2021). uHumans2 is an upgraded version of the uHumans dataset created by MIT/SPARKLAB (Rosinol, Gupta, et al., 2020). The uH2 dataset was generated using Unity and comprises three scenes: a small apartment, an office, and a subway station. It offers visual-inertial data, accurate depth information, and 2D semantic segmentation. Furthermore, the dataset comes with precise robot trajectories, which we used to evaluate our results.

ADE20k (Zhou et al., 2017b) and NYUv2 (Silberman et al., 2012) datasets were used for fine-tuning HRNet and monocular depth networks where Lite-Mono was initialised with ImageNet pre-trained encoder weights.

RGB sequences with IMU data are recorded using the Realsense D435i sensor for the real datasets. RGB sequences are recorded with 848x480 (FOV 69.38 x 42.84 deg) and 640x480 (FOV 55.16 x 42.84 deg) with 30 fps and IMU with 200 Hz. Along with the data sequences, calibration data has been recorded for pre-processing data requirements. The sensor provides depth information, which is not being used and is only recorded in the datasets (ROS Bag files) for comparisons. The recorded data in a building includes observations from five floors, with each floor being individually recorded and the loop between the 2nd and 3rd floors. The data sets are publicly available at https://surfdrive.surf.nl/files/index.php/s/sE0rmSSVQ7wa42a.

**4.1.2 Mono-hydra implementation:** Hydra is implemented on the robotic operating system (ROS), and since Mono-Hydra is an extension of the Hydra framework, deep learning algorithms for 3D mesh generation and VIO is implemented in ROS for easy integration. For Hydra to function properly, it requires odometry to be transmitted as a transformation from a stationary "world" frame to the "sensor_frame" (which is the Realsense camera IMU frame). Consequently, the predicted Depth and odometry data are registered to the camera's IMU frame.

The real-time functionality of Mono-Hydra hinges mainly on the frontend depth (~20-30 fps) and semantic predictions (~10-15 fps), as well as the configuration used to keep the models loaded in CUDA for quicker responses. The semantic segmentation network was the bottleneck among these networks, operating at approximately 15 fps. To maintain synchronisation within the system, the FPS is adjusted accordingly. Ultimately, the Hydra algorithm was set up to utilise a hierarchical descriptor based on 3D scene graphs for loop closure detection.

### 4.2 Results

**4.2.1 Synthetic data - uHumans2 dataset:** Mono-Hydra is tested with a synthetic dataset which is a part of the Hydra experiments, to compare the accuracy of the monocular depth prediction compared to readily available depth in the dataset where the results are tabulated in Table 2. Semantic meshes from the office_s1_ooh dataset are illustrated in Figure 4.

| Scene | DistDepth | | LiteMono | |
|---|---|---|---|---|
| | ME (m) | SD (m) | ME (m) | SD (m) |
| office_s1_ooh | 0.28 | 0.26 | 0.06 | 0.61 |
| apartment_s1_00h | 0.05 | 0.27 | 0.03 | 0.31 |
| subway_s1_ooh | 0.05 | 0.77 | 0.23 | 0.40 |

**Table 2.** Mono-Hydra vs Hydra 3d mesh metric accuracy comparison with two depth prediction networks. ME - Mean Error, SD - Standard Deviation.

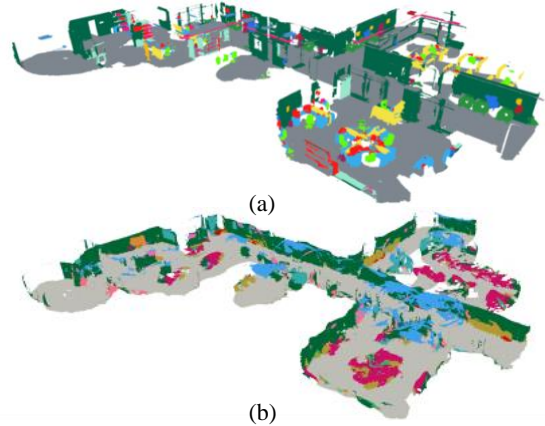

(a)

(b)

**Figure 4.** Semantic meshes[1] generated from (a) Hydra framework (b) Mono-Hydra framework.

---

1. Semantic colour mappings are not the same in uHumans2 dataset and Ade20k dataset.

**4.2.2 Real data – ITC building:** Mono-Hydra is being utilised to generate 3D scene graphs with RGB sequences and IMU data with two different depth-predicting networks, as in Figure 5.

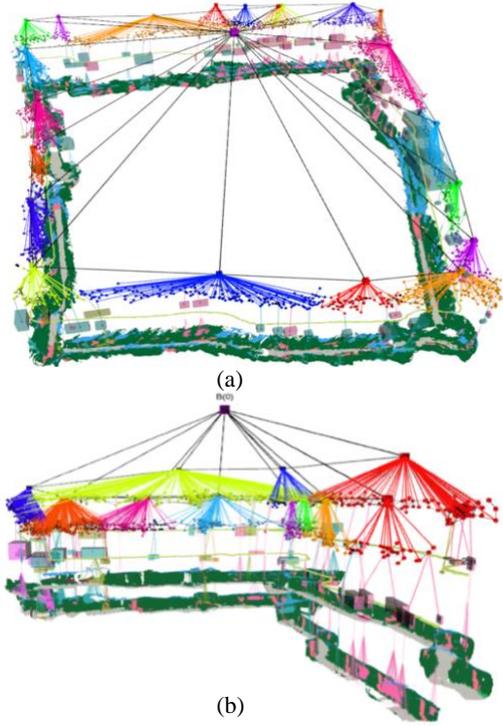

Figure 6 presents the point cloud and semantic mesh for demonstration purposes, where the values are tabulated in Table 4. The distance measurements shown in the figure are examples of how the metric accuracy achieved through monocular depth prediction can be evaluated.

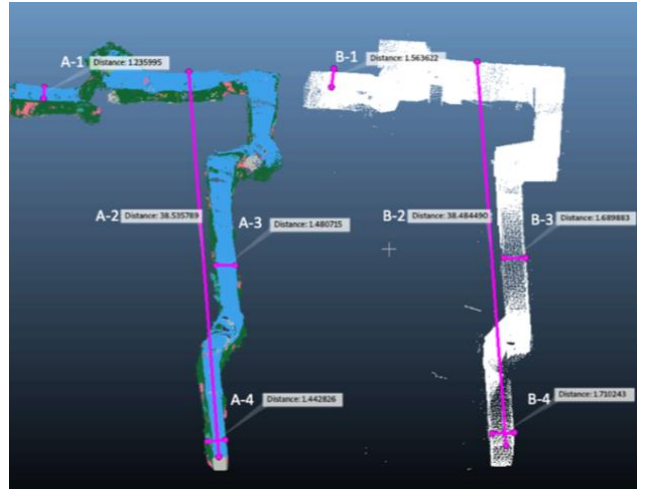

**Figure 6.** Metric results comparison part of the 2nd floor (a) 3d semantic mesh with DistDepth (b) Ground truth.

## 5. DISCUSSION

Our study introduces Mono-Hydra, a significant advancement in robotics' 3D scene understanding, which leverages a monocular camera, IMU data, and real-time algorithms within a highly parallelised architecture.

### 5.1 Methodological Strengths and Limitations

**5.1.1 Synthetic data - uHumans2 dataset:** Mono-Hydra's 3D semantic metric mesh prediction, compared to Hydra, showed variability, as presented in Table 2. While the DistDepth network resulted in an ME ranging from 0.05m to 0.28m, the LiteMono network improved the ME to a range of 0.03m to 0.23m, exhibiting enhanced performance in indoor scenarios. However, LiteMono exhibited higher SD, indicating increased inconsistency, especially notable around the building's ceiling, as illustrated in Figure 7. Lack of sufficient training data, uniformity of ceiling structures, unique capture angles, and varied lighting conditions may have contributed to LiteMono's reduced accuracy in depth predictions near the ceiling.

**Figure 5.** Mono-Hydra results built in real-time with 15 fps on 848 x 480 data sequences (a) Full 2nd floor loop with Lite-Mono (b) 2nd to 3rd floor including stairs with DistDepth.

**4.2.3 Metric-semantic mesh:** The metric accuracy of the generated mesh is compared using CloudCompare (*CloudCompare*, 2023) software to the lidar-based backpack-generated point cloud of the relevant real data (Karam et al., 2019). Firstly, ground truth data is stored as a point cloud, and generated semantic mesh is loaded. The two datasets were then registered using rough and fine registration techniques in CloudCompare software. The results were calculated using the cloud-to-mesh distance option and are presented in Table 3.

| Depth prediction network | 2nd floor | | 3rd floor | |
|---|---|---|---|---|
| | ME (m) | SD (m) | ME (m) | SD (m) |
| DistDepth | 0.19 | 0.18 | 0.21 | 0.16 |
| Lite-Mono | 0.39 | 0.27 | 0.36 | 0.25 |

**Table 3.** Comparison of the metric accuracy of the generated 3d mesh and the ground truth data produced by the 3d lidar backpack. MAE - Mean Error, SD - standard deviation.

| Measurements (m) | | | |
|---|---|---|---|
| 3D semantic mesh | | Ground truth | |
| A-1 | 1.23 | B-1 | 1.56 |
| A-2 | 38.53 | B-2 | 38.48 |
| A-3 | 1.48 | B-3 | 1.69 |
| A-4 | 1.44 | B-4 | 1.71 |

**Table 4.** Selected point comparison in generated 3d mesh and ground truth data in a section of the 2nd floor.

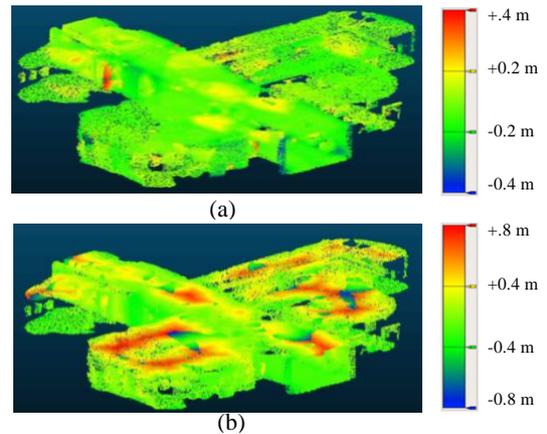

**Figure 5.** Comparison between Hydra vs Mono-Hydra: Metric map of the office (a) DistDepth (b) LiteMono

**5.1.2 Real data - ITC building:** A comparison of the generated mesh accuracy with a lidar-based point cloud reveals that the DistDepth network typically delivers superior accuracy, with a lower Mean Error (ME) between 0.19m to 0.21m as opposed to LiteMono's range of 0.36m to 0.39m. This discrepancy could arise from the suboptimal use of a scaling factor in LiteMono for converting relative depth to metric measurements, compared to DistDepth's direct metric predictions. Differences in network architectures and initial training methodologies also play a part, leading to a two-fold error difference, as illustrated in Table 3. These findings underscore DistDepth's precision in generating a 3D mesh from real-world data.

One notable limitation of our study is the unassessed quality of semantic segmentation in our generated 3D semantic mesh, leaving us without an accuracy measure. Future work should address this, as quantifying the segmentation quality could provide insights into the effectiveness of our approach and indicate areas for further optimisation. This evaluation could enhance Mono-Hydra's overall performance and provide a basis for comparing similar systems, thus contributing to advancements in 3D spatial perception systems.

## 6. CONCLUSIONS

To summarise, this paper introduces Mono-Hydra, a real-time Spatial Perception System that utilises a monocular camera and IMU data to generate a 3D scene graph, achieved by combining real-time algorithms and a highly parallelised perception architecture on a laptop GPU. While the framework shows potential for deployment on embedded systems, further optimisation of deep learning networks is necessary. Moreover, Mono-Hydra provides a persistent representation of the environment with Hydra optimisations and metric-accurate DistDepth monocular depth predictions. The study's results using the DistDepth network, achieving less than 20 cm error in real-time processing, are highly promising. This success is noteworthy considering the monocular camera and real-time setting used. The study demonstrates significant progress in monocular depth prediction and hierarchical metric-semantic mesh generation, showcasing the technology's potential for precise and efficient spatial perception in real-world applications.

Although this approach represents a significant advancement in 3D scene understanding for robotics, there are opportunities for improvement in areas such as hierarchical loop closure optimisation, accurate room detection, and temporal consistency of the generated 3D semantic mesh. Further research should aim to evaluate the multi-task network with temporal consistency, leveraging mutual features in both depth and semantic segmentation and exploring concepts like structure from motion with monocular cameras and deep learning. Furthermore, we have not yet assessed the quality of the semantic segmentation in our generated 3d semantic mesh and thus do not have an accuracy measure for it. This is an important aspect to consider in future work, as it will provide insight into the effectiveness of semantic segmentation and allow for further optimisation of the overall system. In conclusion, the prospective deployment of 3D scene graphs in prediction, planning, and decision-making tasks emerges as a research area that has received comparatively less attention, thereby highlighting a compelling direction for future investigation and progress in this field.